\documentclass[conference]{IEEEtran}
\IEEEoverridecommandlockouts

\usepackage{cite}
\usepackage{tikz}
\usepackage{array}
\usepackage{subfig}     
\usepackage{calc} 
\usepackage{amsmath,amssymb,amsfonts,arydshln}
\usepackage{graphicx}
\usepackage{algorithm}      
\usepackage{algpseudocode}
\usepackage{xspace}
\newcommand{\eg}{e.g.\xspace}

\newcommand{\etal}{et~al.\xspace}
\usepackage{textcomp}
\usepackage{xcolor}
\def\BibTeX{{\rm B\kern-.05em{\sc i\kern-.025em b}\kern-.08em
    T\kern-.1667em\lower.7ex\hbox{E}\kern-.125emX}}
\begin{document}

\title{Causally Guided Gaussian Perturbations for Out-Of-Distribution Generalization in Medical Imaging\\

}

\author{
    \IEEEauthorblockN{
    Haoran Pei\textsuperscript{*, 1},
    Yuguang Yang\textsuperscript{*, 1}\thanks{*These authors have contributed equally and share first authorship.}, 
    Kexin Liu\textsuperscript{1}, 
    Baochang Zhang\textsuperscript{†, 1}\thanks{†Corresponding author : bczhang@buaa.edu.cn}, 
    }
    \\
    \IEEEauthorblockA{\textsuperscript{1}Beihang University, Beijing, China}
}
\maketitle

\IEEEpubidadjcol
\begin{tikzpicture}[remember picture,overlay]
  \node[anchor=south,yshift=10pt] at (current page.south) {\parbox{\textwidth}{\centering\fontsize{7}{8}\selectfont Copyright \copyright\ 2025 IEEE. Published in the Digital Image Computing: Techniques and Applications, 2025 (DICTA 2025), 3--5 December 2025 in Adelaide, South Australia, Australia. Personal use of this material is permitted. However, permission to reprint/republish this material for advertising or promotional purposes or for creating new collective works for resale or redistribution to servers or lists, or to reuse any copyrighted component of this work in other works, must be obtained from the IEEE. Contact: Manager, Copyrights and Permissions / IEEE Service Center / 445 Hoes Lane / P.O. Box 1331 / Piscataway, NJ 08855-1331, USA. Telephone: + Intl. 908-562-3966.}};
\end{tikzpicture}

\begin{abstract}
Out-of-distribution (OOD) generalization remains a central challenge in deploying deep learning models to real-world scenarios, particularly in domains such as biomedical images, where distribution shifts are both subtle and pervasive. While existing methods often pursue domain invariance through complex generative models or adversarial training, these approaches may overlook the underlying causal mechanisms of generalization. 
    In this work, we propose Causally-Guided Gaussian Perturbations (CGP)—a lightweight framework that enhances OOD generalization by injecting spatially varying noise into input images, guided by soft causal masks derived from Vision Transformers. By applying stronger perturbations to background regions and weaker ones to foreground areas, CGP encourages the model to rely on causally relevant features rather than spurious correlations.
   Experimental results on the challenging WILDS benchmark Camelyon17 demonstrate consistent performance gains over state-of-the-art OOD baselines, highlighting the potential of causal perturbation as a tool for reliable and interpretable generalization.
\end{abstract}


\section{Introduction}
In recent years, image classification models have achieved impressive performance on a wide range of benchmarks. However, when deployed in real-world settings where the data distribution may differ from that of the training set—a scenario known as out-of-distribution (OOD) generalization—these models often suffer from significant performance degradation~\cite{liu2021towards,ye2022ood}. This issue is particularly acute in domains such as biomedical imaging~\cite{seyyed2021underdiagnosis}, where subtle shifts in data distribution are common and often unavoidable~\cite{koh2021wilds}. For instance, histopathology slides obtained from different hospitals may differ in staining protocols, scanner types, or sample preparation techniques~\cite{zech2018variable,albadawy2018deep}. Such distributional shifts pose serious challenges to model robustness, as models trained on one domain often fail to generalize reliably to another, even when the underlying classification task remains unchanged.

\begin{figure}[t]
  \centering
  \includegraphics[width=\columnwidth]{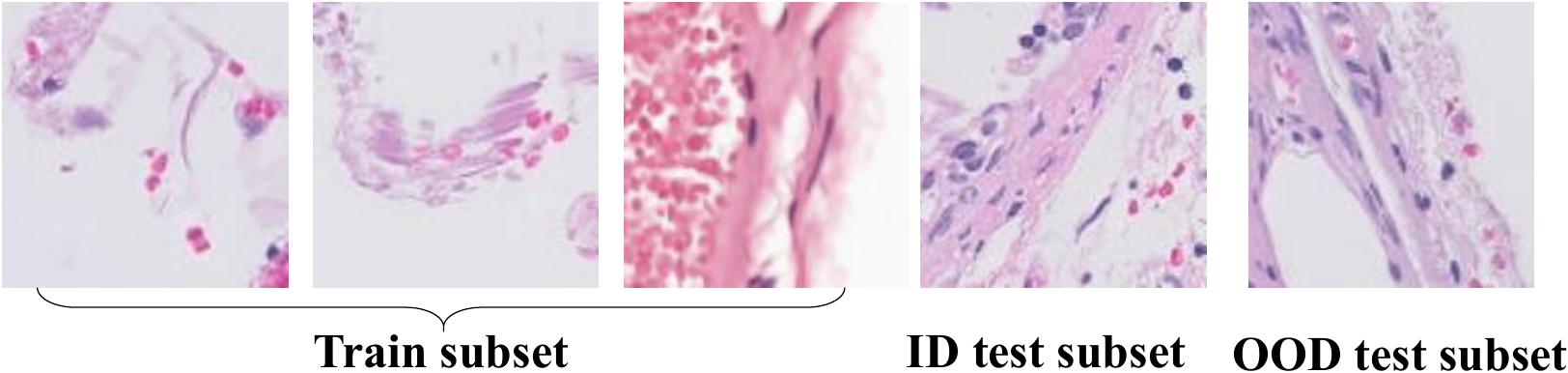} 
  \caption{%
    Five sample patches from the Camelyon17 dataset. Images 1–3 are from the training subset (nodes 0–2); image 4 is from the ID test subset (node 3); image 5 is from the OOD test subset (node 4).}
  \label{fig:samples}
\end{figure}

To address this, a wide array of OOD generalization methods have been proposed. Most of them share a common objective: to learn domain-invariant features that remain stable across environments. Representative techniques include Invariant Risk Minimization (IRM)~\cite{arjovsky2019invariant}, which seeks predictors that perform well across all training environments, and Group Distributionally Robust Optimization (GroupDRO)~\cite{sagawa2019distributionally}, which optimizes for worst-case domain performance. More recent methods, such as V-REx~\cite{krueger2021out} and IRMX~\cite{chen2022pareto}, further refine this idea by reducing variance in risk or penalizing deviations from invariance constraints. These approaches have achieved considerable success and are now widely adopted as baselines in benchmark datasets such as Camelyon17 from the WILDS suite~\cite{koh2021wilds} (see Figure~\ref{fig:samples} for representative samples).

\begin{figure*}[htbp] 
  \centering
  \includegraphics[width=\textwidth]{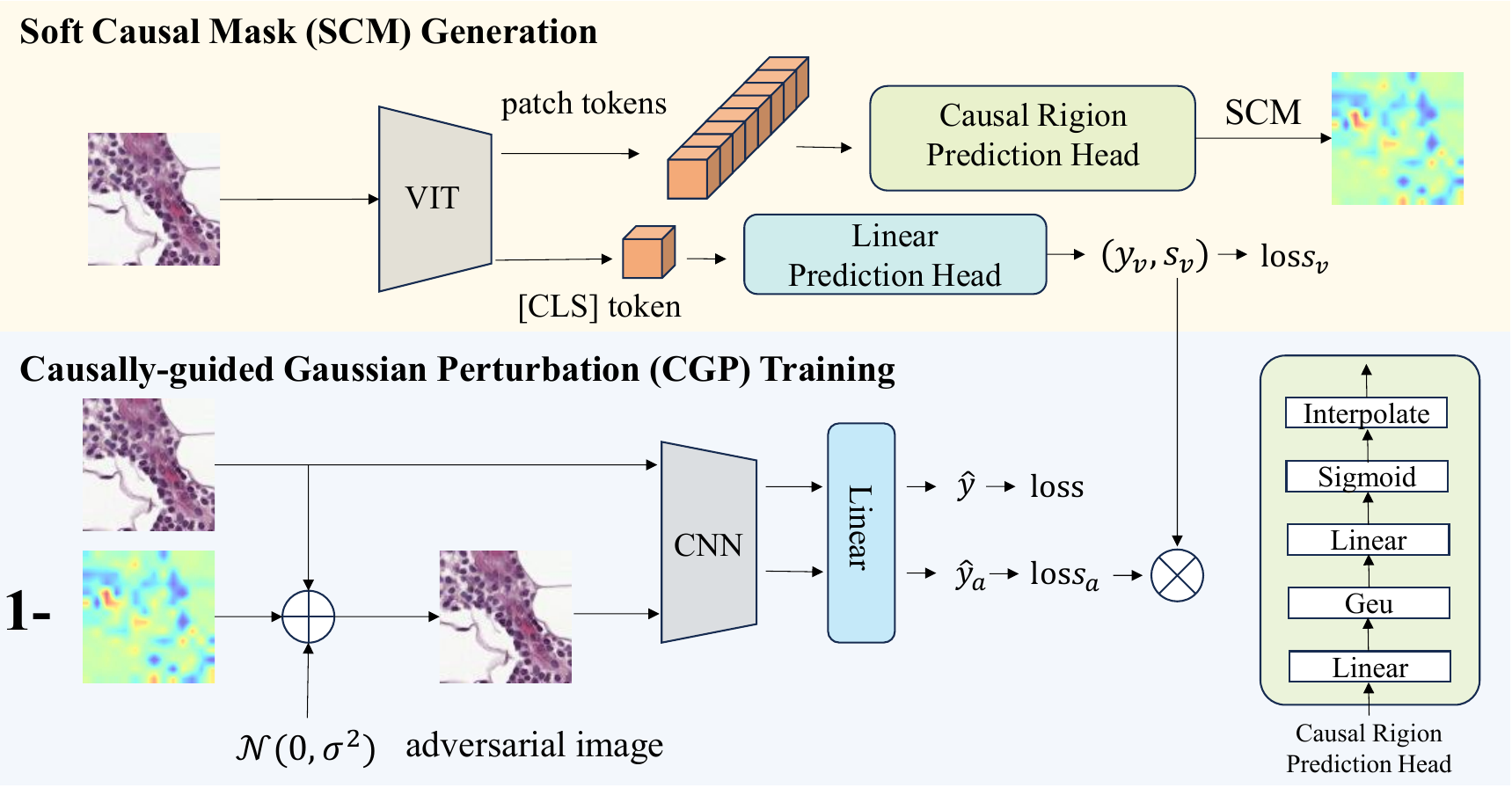}
  \caption{Overview of the proposed CGP framework.
  }
  \label{fig:example}
\end{figure*}

However, these approaches largely treat invariance as a statistical property, without explicitly modeling the causal structure of the input. As a result, the learned invariant features may include contextual patterns that are correlated with the target, such as background textures or co-occurring artifacts, rather than features that are inherently part of the object of interest~\cite{tellez2019quantifying}. This can lead to misleading feature extraction when the scenario is changed~\cite{kamath2021does},~\cite{chen2024diagnosing},~\cite{geirhos2020shortcut}.

This raises a natural question: can we move beyond invariant learning and instead approach OOD generalization from a causal perspective? Motivated by the principle of causal intervention~\cite{wang2022out,wang2022causal}, we introduce a novel framework, Causally-guided Gaussian Perturbation (CGP), that explicitly perturbs image regions identified as influential by the model, thereby encouraging the model to learn the underlying causal structure of the image. 

Specifically, we employ a Vision Transformer (ViT) ~\cite{dosovitskiy2020image} backbone to extract dense patch-token representations of the input image. On top of these features, we introduce a {causal region prediction head} that estimates the spatial distribution of causally relevant regions. The resulting low-resolution causal maps are then upsampled to the input resolution via bilinear interpolation to produce a full-resolution Soft Causal Mask (SCM). Then, the resulting SCM is used to guide the application of spatially varying Gaussian perturbations on the input image. In this way, stronger noise is applied to background regions (low SCM values), while foreground regions (high SCM values) receive only mild perturbations. This asymmetric masking-perturbation process serves as a causal intervention procedure: by selectively disrupting non-causal or spurious regions, CGP encourages the model to rely on invariant, causally relevant features that can generalize across domains. Furthermore, since the causal region prediction head may not yield reliable SCM masks during early training, we introduce a confidence-based weighting strategy to stabilize optimization. Specifically, we leverage the ViT’s [CLS] token representation to obtain a confidence score (see Fig.~\ref{fig:example}). This score is used to modulate the loss contribution from adversarial (perturbed) samples: samples with higher classification confidence receive greater weight, while uncertain samples exert less influence. This encourages the model to gradually learn from more reliable supervision while the SCM predictions mature.

In contrast to previous methods that rely on generative models~\cite{ktena2024generative} or adversarial attacks~\cite{yi2021improved} for data augmentation, our approach is directly grounded in causal learning principles—modifying causal regions to reveal invariant predictive mechanisms. 

We validate our method on the Camelyon17 dataset. Extensive experiments show that our model consistently outperforms strong baselines, including IRM, GroupDRO, V-REx, and IRMX, achieving an average accuracy gain of 1\%-5\% on OOD test domains. Our contributions are as follows:
\begin{itemize}
    \item {We propose a new causal intervention framework for OOD generalization in image classification, built upon vision transformers and mask-guided perturbations.}
    \item {We demonstrate that simple Gaussian noise, when used in a causally structured manner, can serve as an effective alternative to complex generative modeling approaches.}
    \item {We conduct comprehensive evaluations on a real-world dataset, achieving superior OOD performance compared to the state-of-the-art baseline.}
\end{itemize}

\section{Related Work}
\textbf{Vision Transformers:} ViT is a transformer-based architecture for image recognition tasks, which processes an input image by dividing it into fixed-size patches, embedding them linearly and feeding the resulting sequence into a standard Transformer encoder. Using self-attention mechanisms, ViT can effectively model long-range dependencies and capture global contextual information across the entire image~\cite{khan2022transformers,han2022survey}. Compared with convolutional neural networks (CNNs), which rely on local receptive fields, ViT demonstrates stronger feature representation capabilities and has been successfully applied to a wide range of computer vision tasks such as image classification~\cite{dosovitskiy2020image}, object detection~\cite{carion2020end}, and semantic segmentation~\cite{zheng2021rethinking}. Recent studies have increasingly explored the generalizability of ViT, especially in OOD settings~\cite{zhang2022delving,sultana2022self}. Unlike CNNs, which tend to rely heavily on local textures and background cues, ViTs focus more on shape and structural information, aligning more closely with the human visual system. This intrinsic difference enables ViTs to exhibit better robustness and generalization under distributional shifts, making them particularly suitable for tasks involving spurious correlation and biased training data.

\textbf{Weakly Supervised Semantic Segmentation:} Weakly Supervised Semantic Segmentation (WSSS) aims to train segmentation models using weak annotations such as image-level labels, bounding boxes, or scribbles, instead of costly pixel-level masks~\cite{zhang2020survey}. Among these, image-level labels are the most economical but also the most ambiguous, posing significant challenges for accurate localization.To address this, many approaches use Class Activation Maps (CAM)  to generate initial localization cues~\cite{ahn2018learning,10.1007/978-3-319-46493-0_42}, which are then refined through strategies including saliency guidance~\cite{Khoreva_2017_CVPR}. Recent works also explore using vision transformers to enhance spatial coherence and global context modeling in WSSS pipelines~\cite{rossetti2022max}. Despite their progress, WSSS methods often struggle with background confusion and incomplete object coverage, motivating further research into more robust supervision signals.

\textbf{Causal Inference: }Recent studies have demonstrated that causal inference techniques can effectively mitigate spurious correlations such as background bias in image classification tasks~\cite{mao2021cgen,venkataramani2024causal}. Conventional deep learning models often capture non-causal features that are spuriously associated with the label (\eg textures or co-occurring backgrounds), which can lead to significant performance degradation when the test environment changes~\cite{beery2018recognition,arjovsky2019invariant}. To address this, researchers have proposed causal intervention methods to uncover stable, invariant features and suppress spurious correlations.Wang \etal~ \cite{wang2021caam} proposed the Causal Attention Module (CAM), which identifies and masks contextual distractions in a weakly supervised manner, thereby forcing the model to focus on foreground causal cues. Similarly, Mao \etal~ \cite{mao2021cgen} introduced a generative intervention framework that synthesizes images with diverse backgrounds, enabling the model to disentangle and remove confounding factors. Venkataramani \etal~ \cite{venkataramani2024causal} further advanced this direction by proposing Causal Feature Alignment (CFA), which uses limited segmentation supervision to align foreground features while suppressing background bias. Their method achieves state-of-the-art performance on challenging benchmarks such as Waterbirds and the ImageNet-9 background shift dataset. Causal intervention approaches explicitly model causal structures, thus demonstrating superior generalization under noisy or shifted conditions.

\section{Methodology}

We propose the CGP module as a general, plug-and-play enhancement strategy for deep visual classifiers. This module can be seamlessly integrated into various training frameworks. The structure of the proposed module is depicted in Figure~\ref{fig:example}. In this work, we instantiate it on the Feature Augmented Training(FeAT)~\cite{chen2023understanding} algorithm due to its strong performance on domain generalization benchmarks, though our method is not limited to FeAT and can be readily extended to other backbones.

The core idea is to leverage a ViT, which can be either pretrained or jointly trained with the main model, to generate spatial masks that highlight semantically critical regions in the input image. These masks guide a causal perturbation mechanism that applies soft interventions by injecting noise into less discriminative regions, thereby encouraging the model to concentrate on causally relevant features while being robust to spurious correlations.

\subsection{ViT-based Soft Causal Mask (SCM) }

The soft causal mask is generated by feeding the input image $x$ into the ViT encoder to obtain patch tokens. These tokens are then passed through a linear layer followed by a sigmoid activation, producing soft attention scores for each patch. The resulting scores are upsampled via bilinear interpolation to match the original image resolution, yielding a continuous-valued mask $M \in [0, 1]^{H \times W}$. This allows for fine-grained and learnable control over the perturbation regions, rather than relying on manually designed thresholds or binary selections. Formally:

\begin{equation}
M = \text{Upsample}\left( \sigma(W_{\text{mask}} \cdot T) \right),
\end{equation}

where $T$ denotes patch tokens' {embedding} from the ViT encoder, $W_{\text{mask}}$ is a learnable projection, and $\sigma(\cdot)$ is the sigmoid function. The soft mask emphasizes causally important regions while assigning lower values to potentially spurious or context-dependent patches.

\subsection{Noise Injection Strategy}

To perturb the less discriminative regions, we inject Gaussian noise into the input image based on the generated mask. Specifically, the perturbed image is computed as:

\begin{equation}
x_{\text{adv}} = x \cdot M + (1 - M) \cdot \mathcal{N}(0, \sigma^2),
\end{equation}

where $\mathcal{N}(0, \sigma^2)$ represents Gaussian noise with standard deviation $\sigma$. Unlike feature-level perturbations that rely on the assumption of feature independence—which is often violated in CNNs due to strong channel coupling—pixel-level SCM guided noise injection avoids this issue and preserves the interpretability of perturbation regions.

\subsection{Adaptive Adversarial Weighting}

We introduce an input-dependent adversarial loss weighting function $\lambda_{\text{adv}}(x)$ based on the ViT’s prediction confidence. A custom sigmoid-based weighting function ensures that only sufficiently confident ViT outputs contribute significantly to the adversarial training loss:

\begin{equation}
\lambda_{\text{adv}}(x) = \tau + (1 - \tau) \cdot \left( \frac{1}{1 + e^{-k (c(x) - \tau)}} - 0.5 \right) \cdot 2,
\end{equation}

where $c(x)$ is the softmax confidence of the ViT prediction, $\tau$ is the target threshold (\eg 0.75), and $k$ controls the steepness. This formulation ensures smooth weighting while suppressing the influence of noisy or uncertain ViT outputs.

\subsection{Training Objective and Two-stage Strategy}

Our training objective integrates three components: standard classification loss $\mathcal{L}_{\text{orig}}$ on clean inputs, ViT-guided auxiliary loss $\mathcal{L}_{\text{vit}}$, and perturbation-aware loss $\mathcal{L}_{\text{adv}}$:

\begin{equation}
\mathcal{L}_{\text{total}} = \mathcal{L}_{\text{orig}} + \lambda_{\text{vit}} \cdot \mathcal{L}_{\text{vit}} + \lambda_{\text{adv}}(x) \cdot \mathcal{L}_{\text{adv}},
\end{equation}

where all losses are cross-entropy losses computed from the respective predictions. This design encourages the model to focus on causally relevant regions while remaining robust to adversarial noise.

Moreover, to improve efficiency and stabilize training, we adopt a two-stage training pipeline: in the first stage, both ViT and perturbation modules are used to guide feature learning; in the second stage, we discard the ViT and perturbation modules and fine-tune only the CNN-based classifier on clean inputs. This not only improves generalization but also ensures that inference is efficient and ViT-free.

\subsection{Training Workflow}

The overall training workflow is summarized below:

\begin{algorithm}[H]
\caption{Training Workflow of CGP}
\label{alg:vit_training}
\textbf{Input:} Training data $X$ with labels $y$; ViT model; Gaussian noise strength $\sigma$; confidence threshold $\tau$; feature extractor $f(\cdot)$; classifier $h(\cdot)$; number of iterations $K$ \\
\textbf{Output:} Trained model $h(f(\cdot))$
\begin{algorithmic}[1]
\For{$k = 1$ to $K$}
    \State Obtain patch tokens: $T = \text{ViT}_{\text{enc}}(X)$
    \State Generate soft mask: $M = \text{Upsample}(\sigma(W_{\text{mask}} \cdot T))$
    \State Compute perturbed inputs: $X_{\text{adv}} = X \cdot M + (1 - M) \cdot \mathcal{N}(0, \sigma^2)$
    \State Extract features: $F = f(X)$; $F_{\text{adv}} = f(X_{\text{adv}})$
    \State Predict labels: $\hat{y} = h(F)$; $\hat{y}_{\text{vit}} = h(f(M \cdot X))$; $\hat{y}_{\text{adv}} = h(F_{\text{adv}})$
    \State Compute loss: \[
        \mathcal{L} = \text{CE}(\hat{y}, y) + \lambda_{\text{vit}} \cdot \text{CE}(\hat{y}_{\text{vit}}, y) + \lambda_{\text{adv}}(x) \cdot \text{CE}(\hat{y}_{\text{adv}}, y)
    \]
    \State Update model parameters
\EndFor
\State Fine-tune classifier $h(f(\cdot))$ on clean inputs $X$
\State \Return $h(f(\cdot))$
\end{algorithmic}
\end{algorithm}

\section{EXPERIMENTAL SETUP}

\subsection{Dataset}
We conduct experiments on the \textbf{Camelyon17} dataset to evaluate the robustness and domain generalization ability of our method under realistic distribution shifts. This dataset is provided as part of the WILDS benchmark suite \cite{koh2021wilds}, which focuses on evaluating models under real-world domain shifts. We follow the official training, validation, and test splits defined by WILDS to ensure fair comparison with prior work and maintain experimental consistency.

\textbf{Camelyon17} is a large-scale binary classification dataset derived from the CAMELYON17 Grand Challenge. It contains high-resolution histopathology whole-slide images (WSIs) of lymph node tissue, labeled to indicate the presence or absence of metastatic cancer. The WSIs are collected from five independent hospitals, each considered a distinct domain, thus providing a natural source of distribution shift due to variations in patient demographics, tissue preparation protocols, slide scanning equipment, and staining techniques.

From these WSIs, over 450{,}000 tissue \emph{patches} of size $224 \times 224$ pixels are extracted and labeled accordingly. Specifically, the training set includes 302{,}436 patches sampled from three hospitals (domains 0–2). The validation set consists of 34{,}038 samples from a fourth hospital (domain 3), which serves as the in-domain (ID) validation set for assessing generalization within the distribution of training domains. The test set contains 119{,}480 samples from a fifth hospital (domain 4), which constitutes the OOD test set, used to evaluate model performance under unseen distributional shifts. No labels or domain identifiers from the validation or test hospitals are used during training, strictly enforcing an OOD evaluation protocol.

Camelyon17 poses several challenges for domain generalization due to the heterogeneous nature of histopathology data. Domain shifts manifest in the form of color variation from different staining procedures, structural differences in tissue morphology, and differences in scanner resolution and lighting conditions. These factors make the dataset a strong benchmark for evaluating a model's ability to generalize across clinical settings. Additionally, the large scale of the dataset introduces further complexity: with over 450,000 labeled patches extracted from whole-slide images, Camelyon17 is significantly larger than typical medical imaging datasets. This not only imposes a high computational and storage burden but also increases the difficulty of training due to class imbalance, patch-level noise, and subtle inter-class variation. The high-resolution nature of the images and the diversity of acquisition conditions make it difficult for models to capture domain-invariant features without overfitting to source-specific patterns.

Consequently, Camelyon17 has been widely adopted in the literature to benchmark algorithms for DRO, domain adaptation, and OOD generalization. By focusing on Camelyon17, we specifically target the real-world challenge of deploying machine learning models across multiple hospitals or healthcare providers where data distribution may vary significantly. Our experimental design emphasizes not only overall classification performance, but also robustness to domain shift, which is essential for the safe deployment of AI systems in biomedical contexts.

\subsection{Experimental Details}
We adopt \textbf{FeAT} as our primary baseline. It partitions the training data into retention and augmentation sets to enhance feature diversity. We additionally compare with the standard \textbf{ERM} strategy to evaluate performance gains over traditional training paradigms. Our proposed method integrates into FeAT without altering its core pipeline. All hyperparameters, including optimizer, learning rate schedule, number of epochs, and augmentation strategies, are kept consistent with the official FeAT implementation for fair comparison.

\textbf{Data preprocessing:} All Camelyon17 images are resized to $224 \times 224$, converted to tensors, and normalized using the standard ImageNet mean and standard deviation: $[0.485, 0.456, 0.406]$ and $[0.229, 0.224, 0.225]$, respectively. During training, we apply random 90-degree rotations and horizontal flips. During evaluation, only normalization and tensor conversion are applied.

\textbf{Model and ViT configuration:} To enable spatially-aware interventions, we incorporate a ViT module to generate attention masks. The ViT model is initialized from the Hugging Face Transformers library, pretrained on ImageNet-21k and fine-tuned on ImageNet-1k. For Camelyon17, we train the ViT module from scratch to ensure fair comparison with the from-scratch DenseNet-121 encoder used in FeAT. Bilinear interpolation is used to adapt input images and positional embeddings to the required patch size. The classification head is replaced to match the binary task, and all ViT layers are trained end-to-end.

\textbf{Training setup:} We use the same number of training epochs as FeAT. The batch size is set to 32. The ViT attention loss is weighted by $\lambda_{\mathrm{vit}} = 0.1$. A confidence function with threshold 0.75 and steepness parameter 8 is employed to suppress noisy gradients from low-confidence masks, enhancing robustness. Learning rate and warm-up schedules are retained as in the FeAT baseline.

\textbf{Mask Visualization Analysis:} We qualitatively evaluate the effectiveness of the attention masks by comparing \textbf{ViT-generated} masks against standard \textbf{CAM } heatmaps. Both visualization techniques are applied to high-confidence test images. We observe that ViT-based masks consistently highlight more semantically meaningful regions associated with cancerous tissue, while CAM often produces broader and less focused attention regions. This analysis supports our claim that ViT-guided interventions focus the model's learning on more relevant spatial features, contributing to improved generalization.

\textbf{Evaluation:} 
We report classification accuracy on both the in-domain (ID) validation set and the out-of-domain (OOD) test set. These metrics allow us to examine whether our method improves OOD generalization while maintaining ID performance. All evaluations are conducted using the same data splits and model checkpoints as the FeAT and ERM baselines to ensure comparability.

\subsection{Hardware and Environment}
All experiments are conducted on a single NVIDIA RTX 4090 GPU with 48 GB of VRAM. The software environment includes Ubuntu 22.04, PyTorch 1.11.0, and the Hugging Face Transformers library. We run each experiment with ten different random seeds and report average performance to reduce variance and ensure statistical robustness.

\newlength{\picW}
\setlength{\picW}{\linewidth/6}   

\begin{figure*}[t!]
  \centering
  \renewcommand\arraystretch{0.8}

  \begin{tabular}{@{} *6{c} @{}}
    \includegraphics[width=\picW]{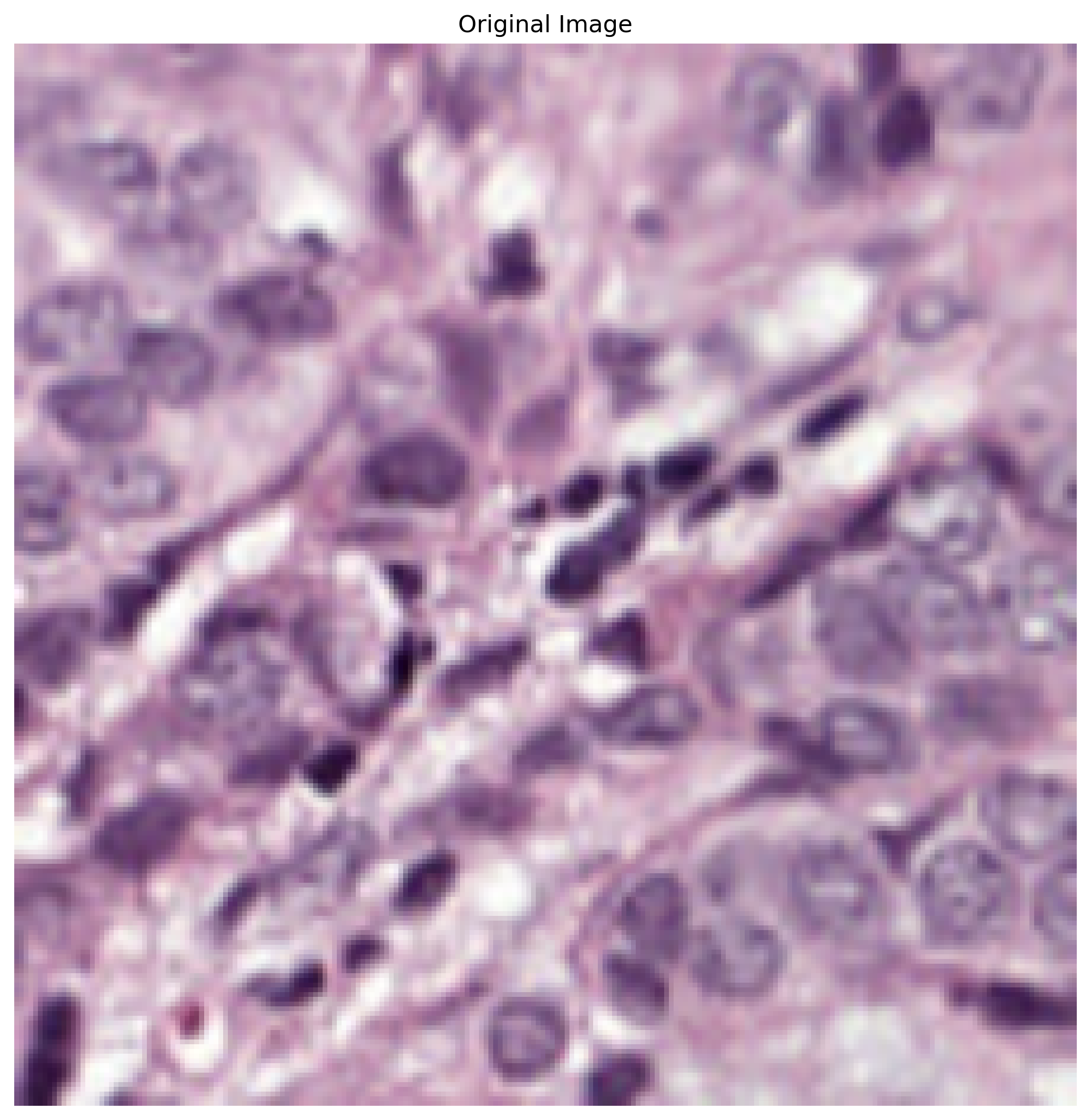} &
    \includegraphics[width=\picW]{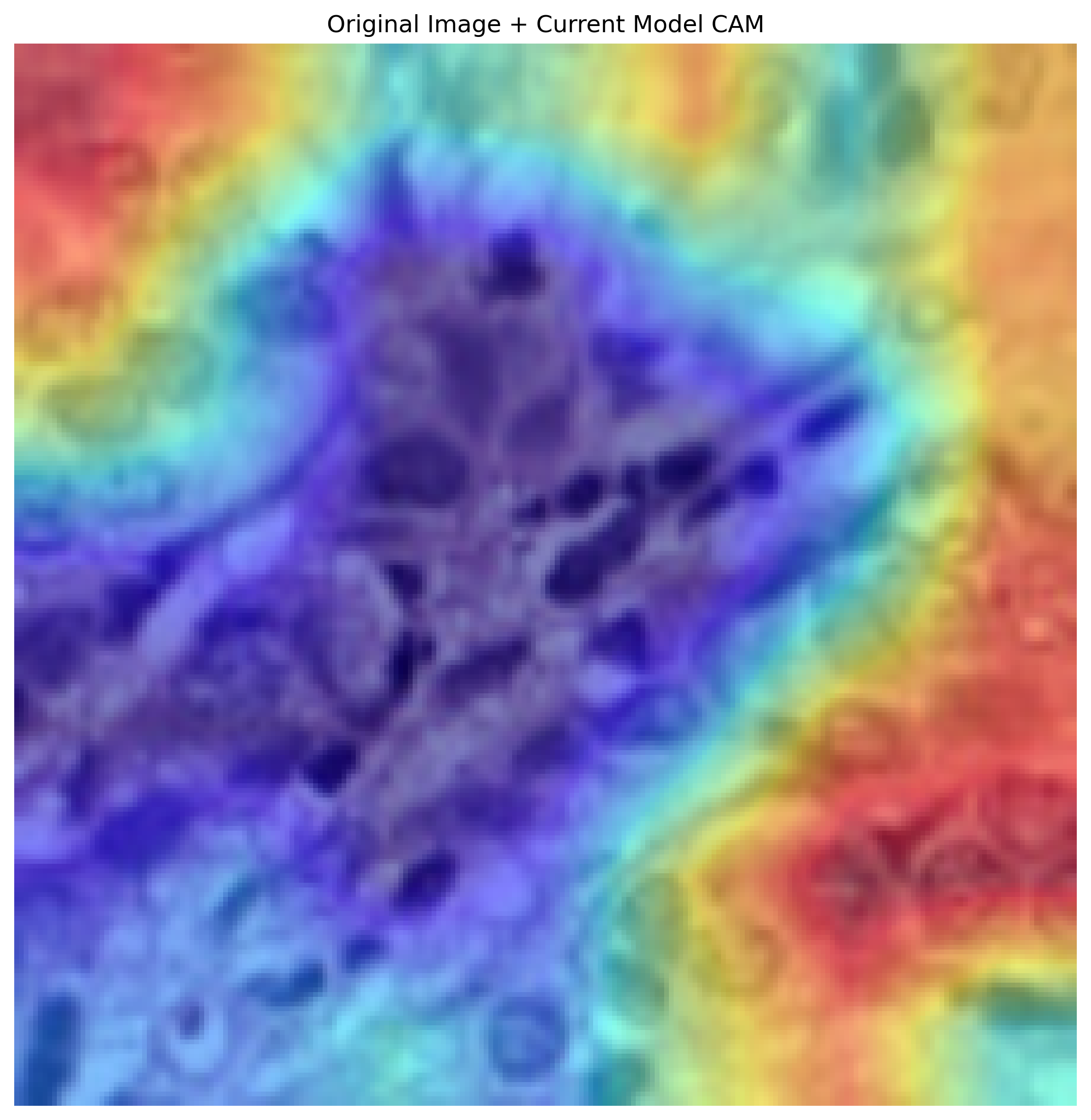} &
    \includegraphics[width=\picW]{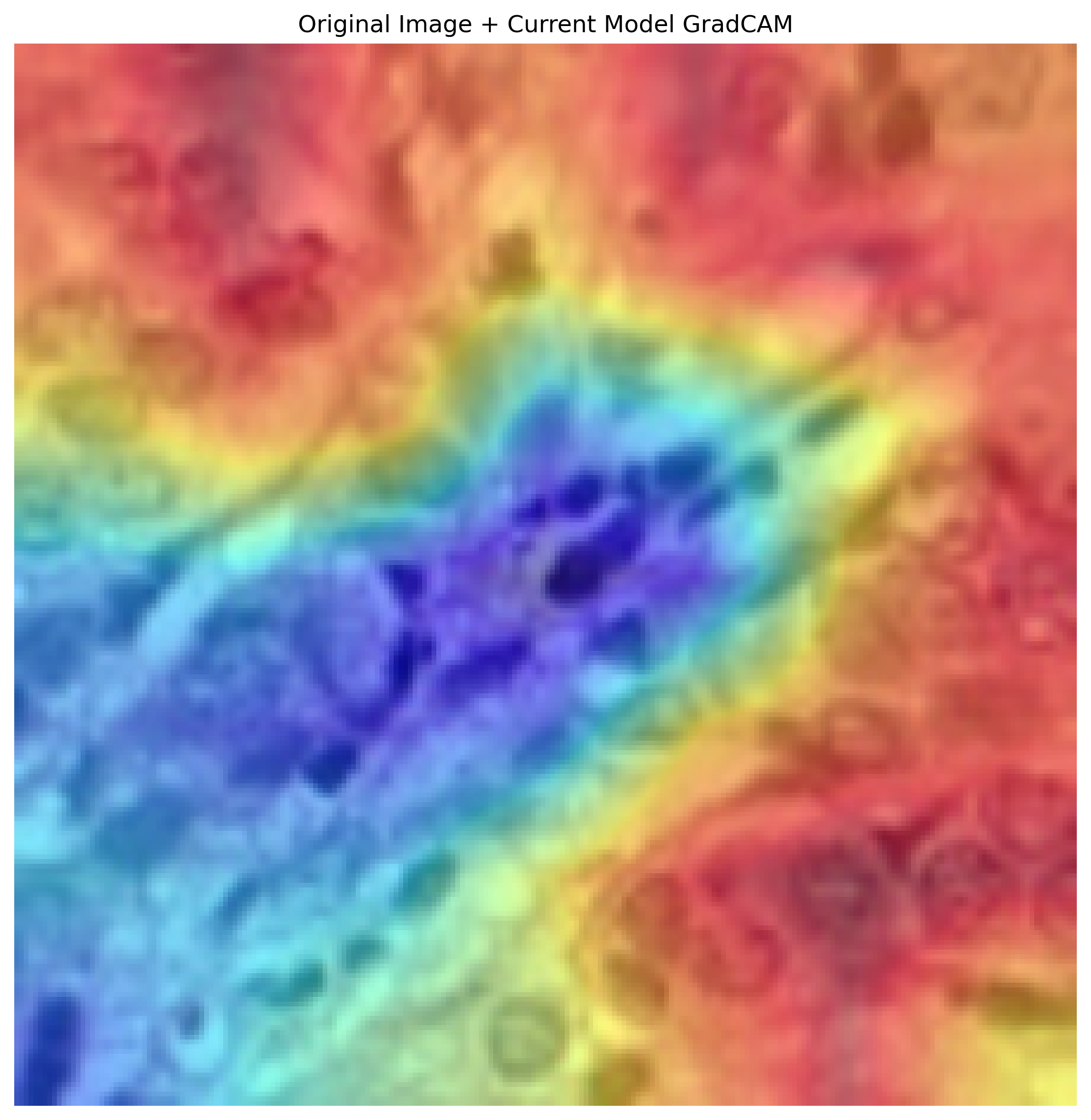} &
    \includegraphics[width=\picW]{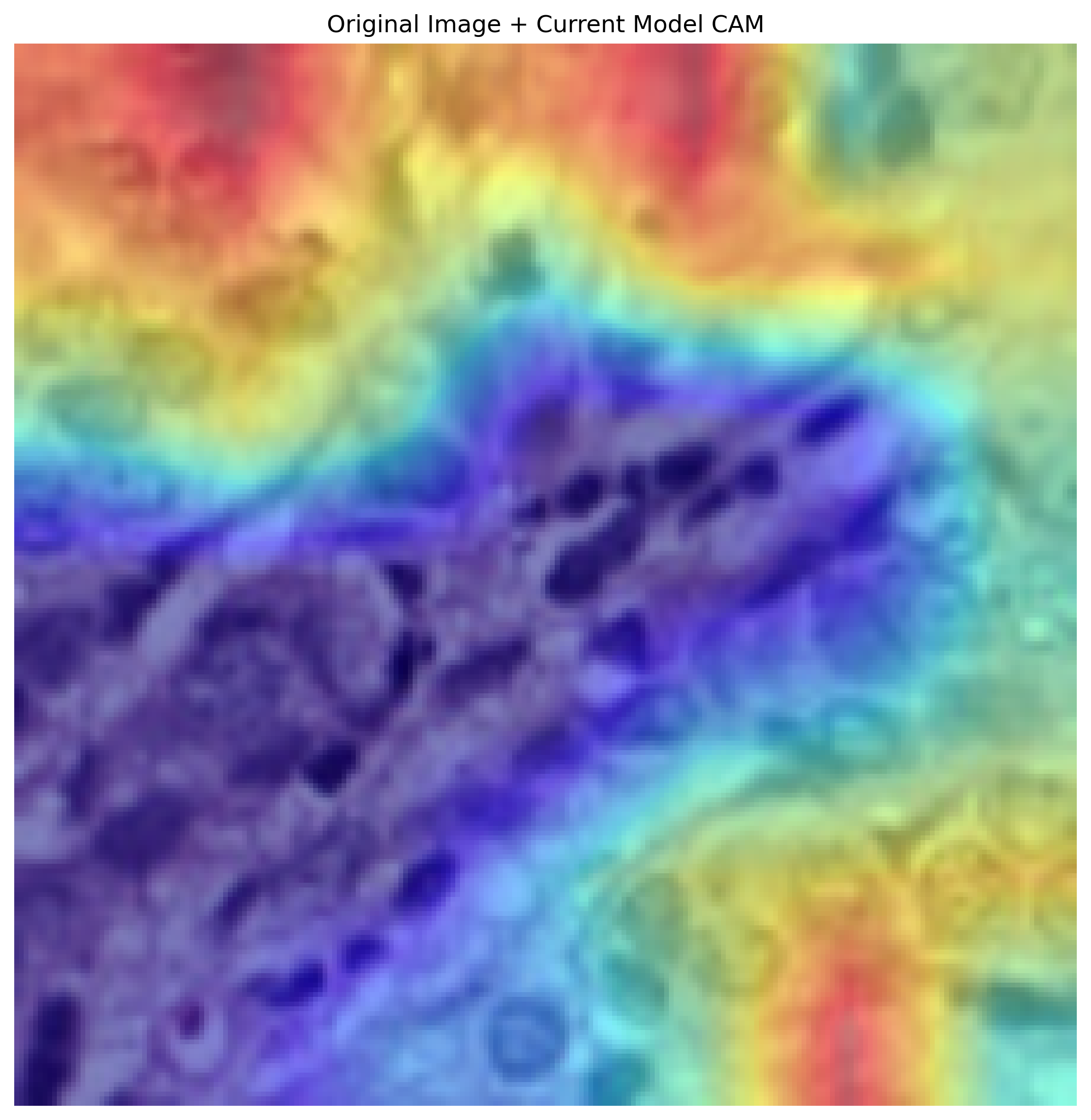} &
    \includegraphics[width=\picW]{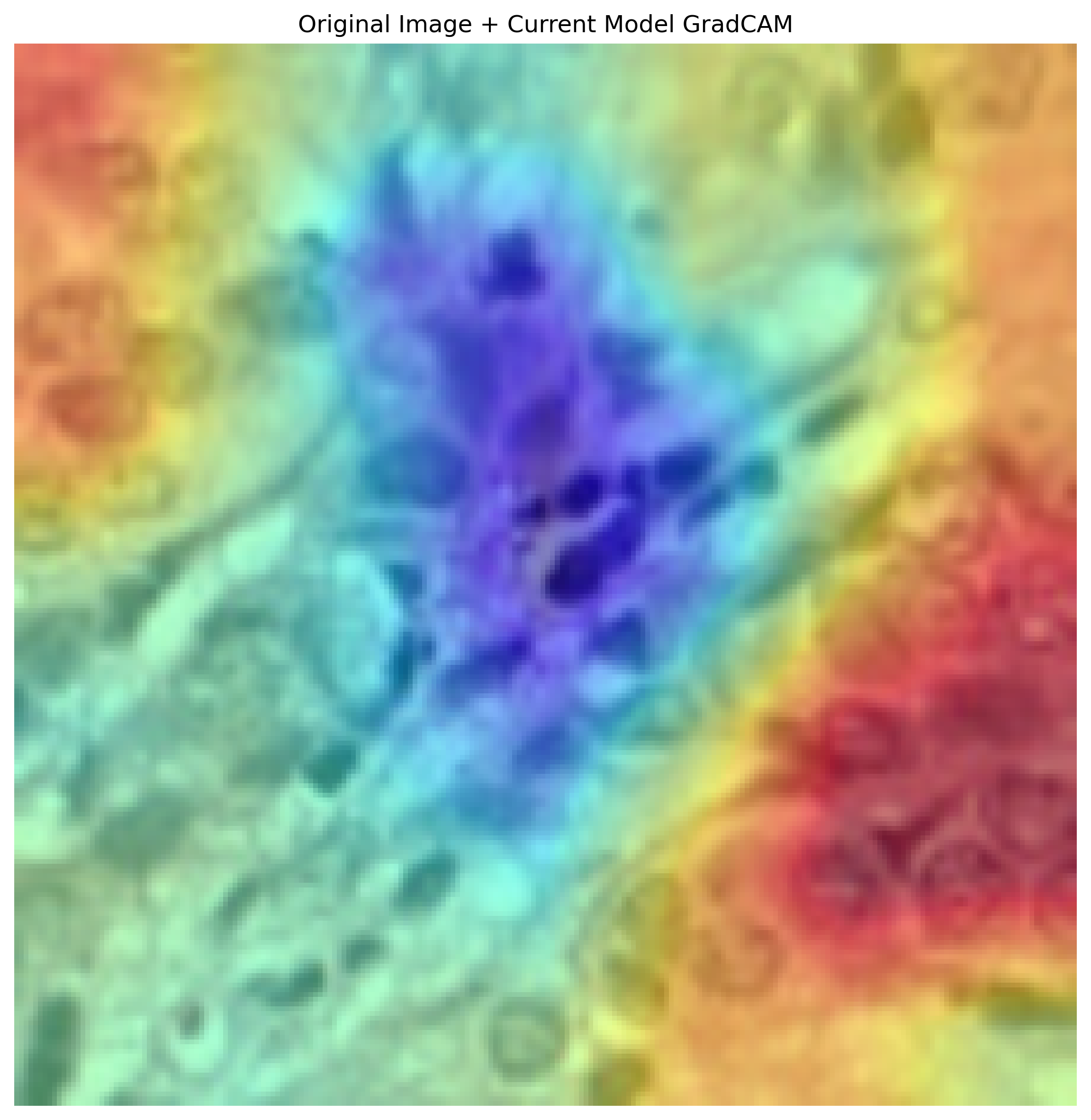} &
    \includegraphics[width=\picW]{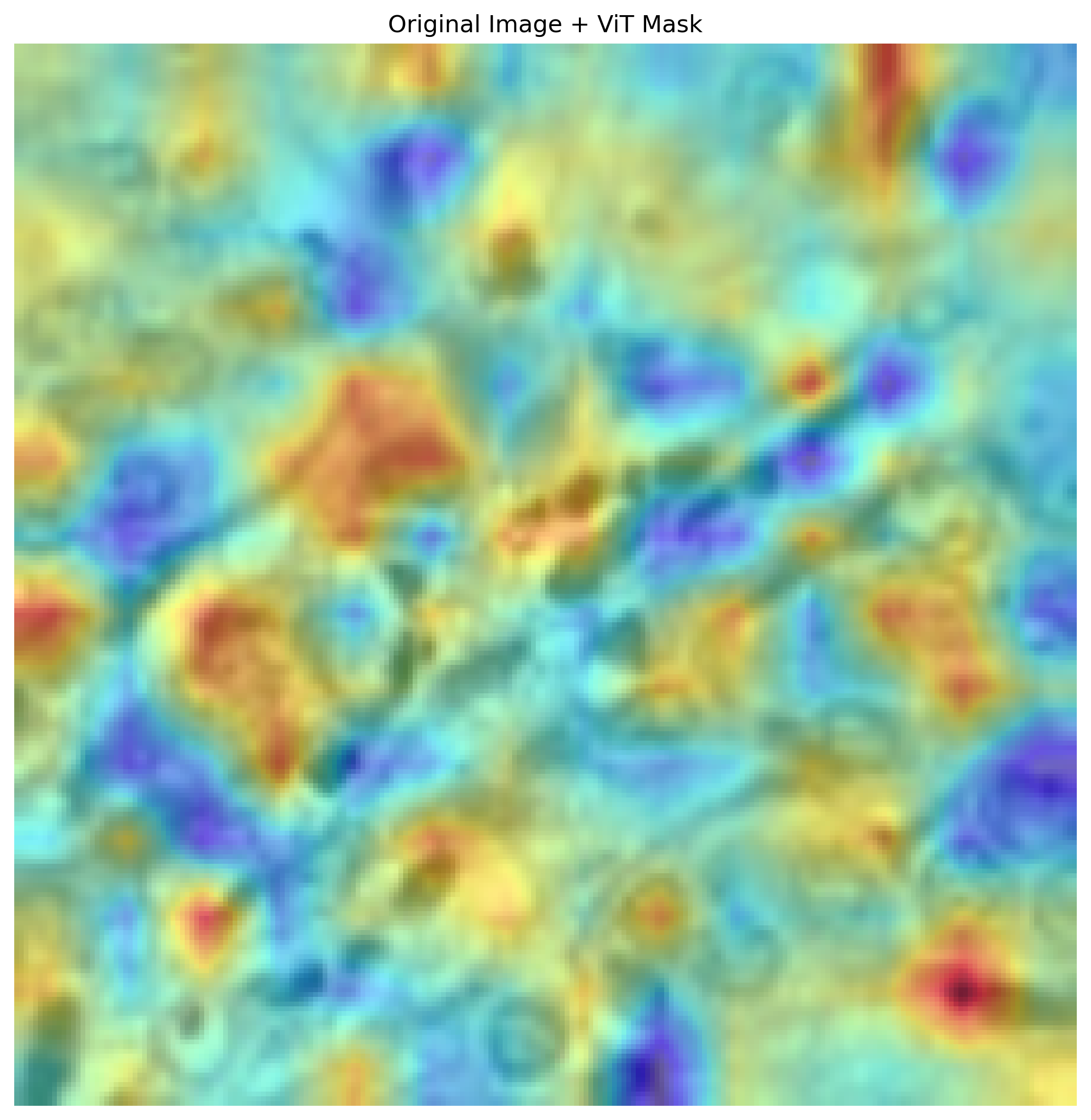} \\[-0.5ex]

    \includegraphics[width=\picW]{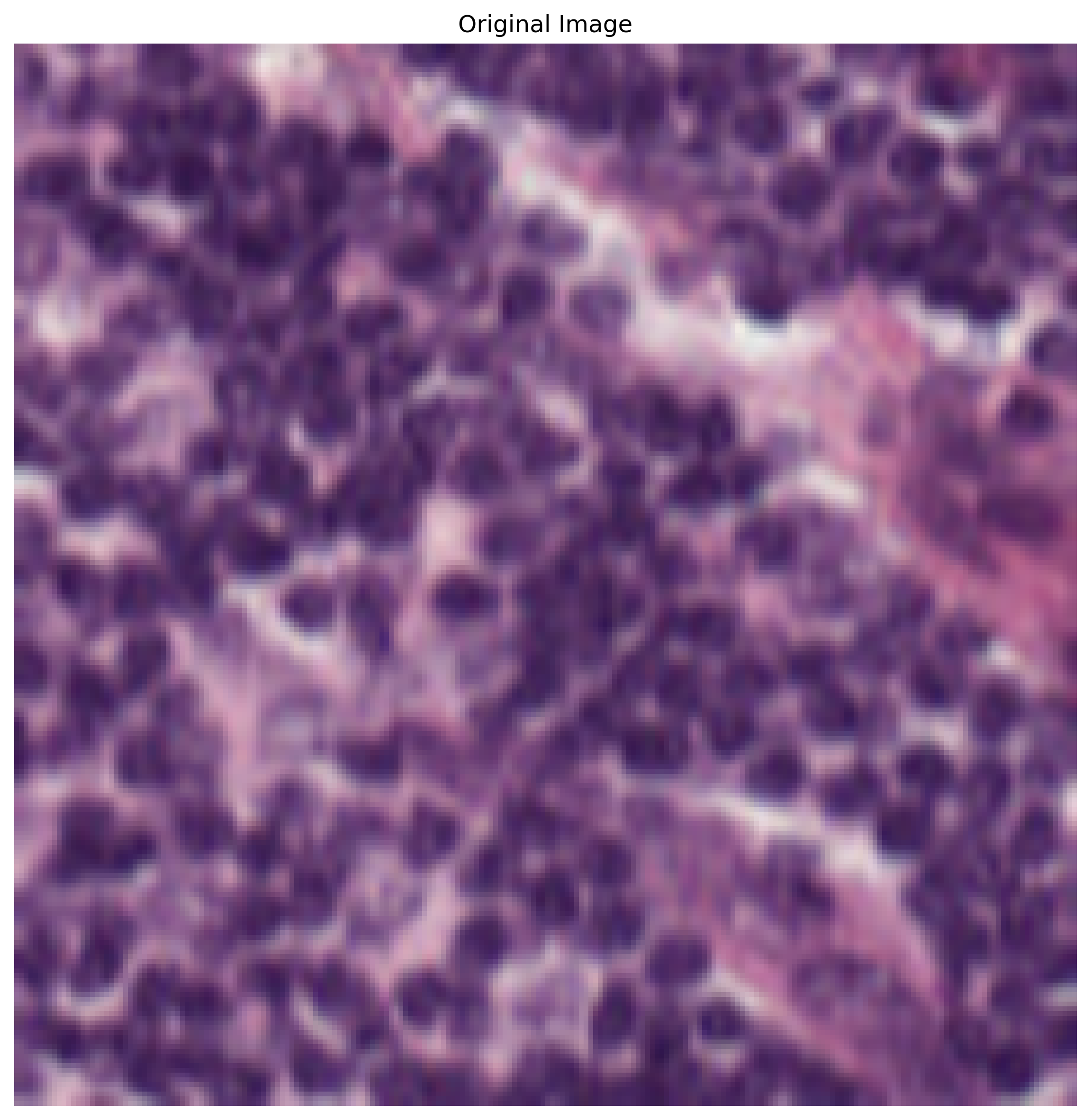} &
    \includegraphics[width=\picW]{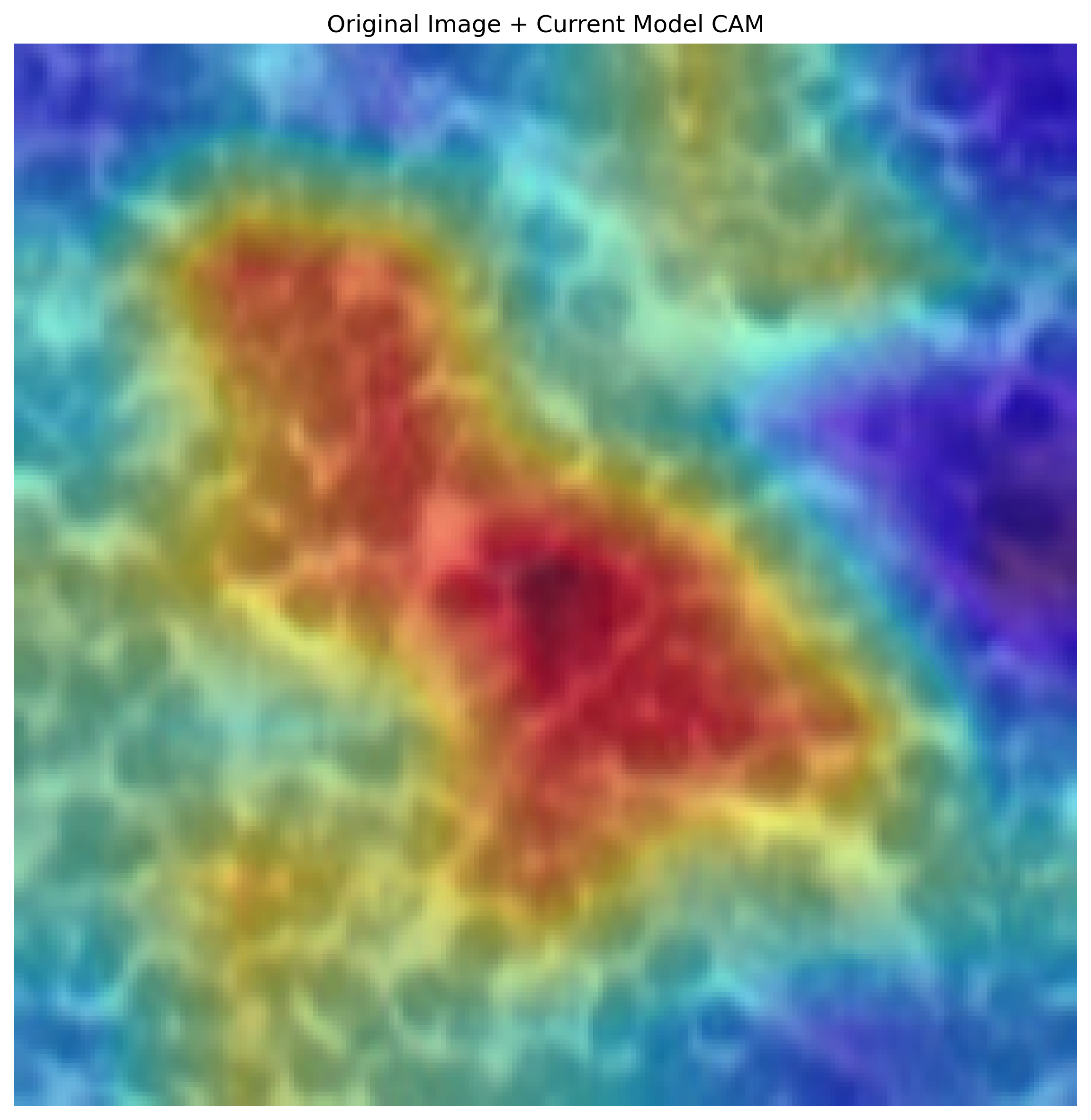} &
    \includegraphics[width=\picW]{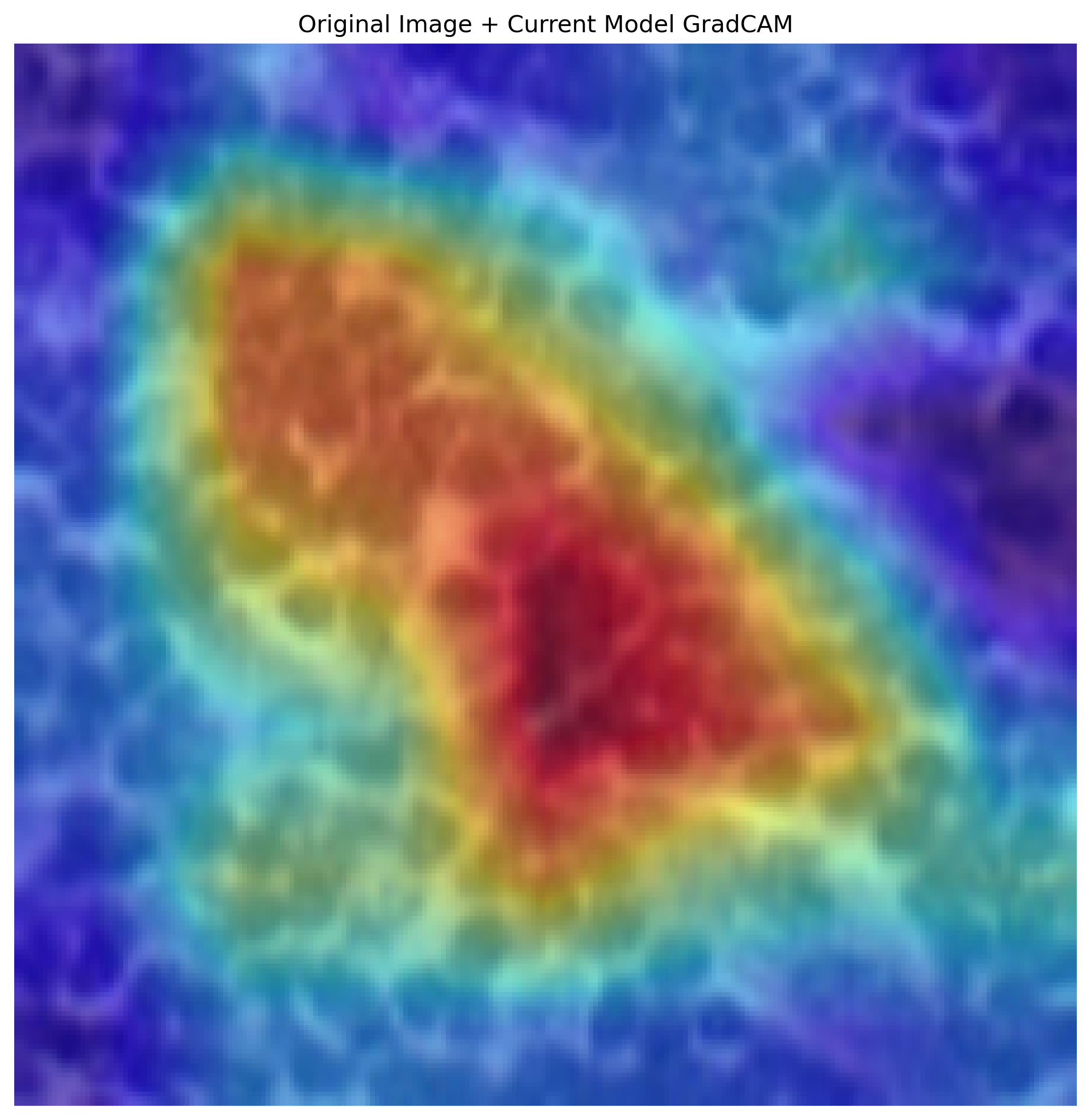} &
    \includegraphics[width=\picW]{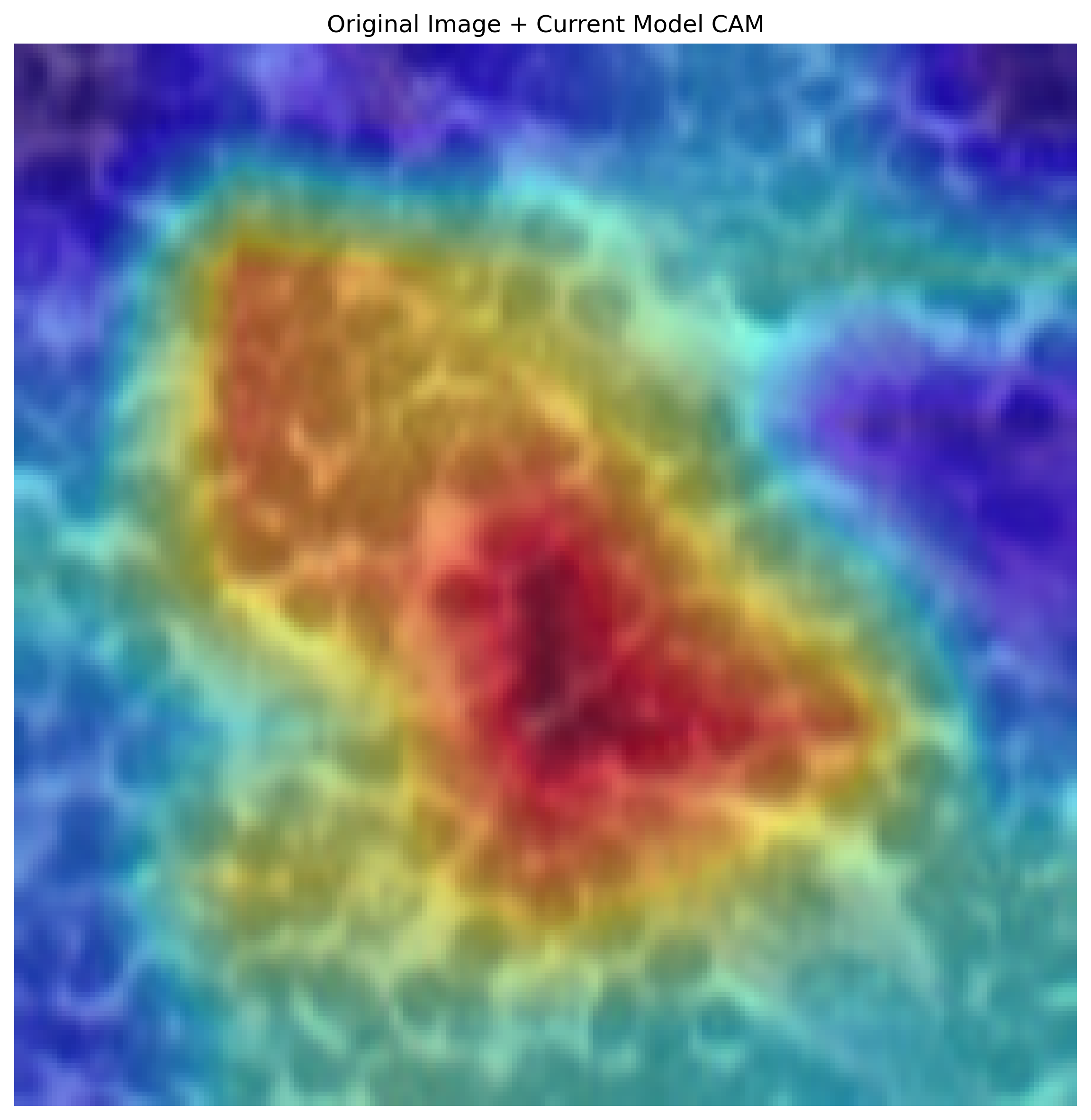} &
    \includegraphics[width=\picW]{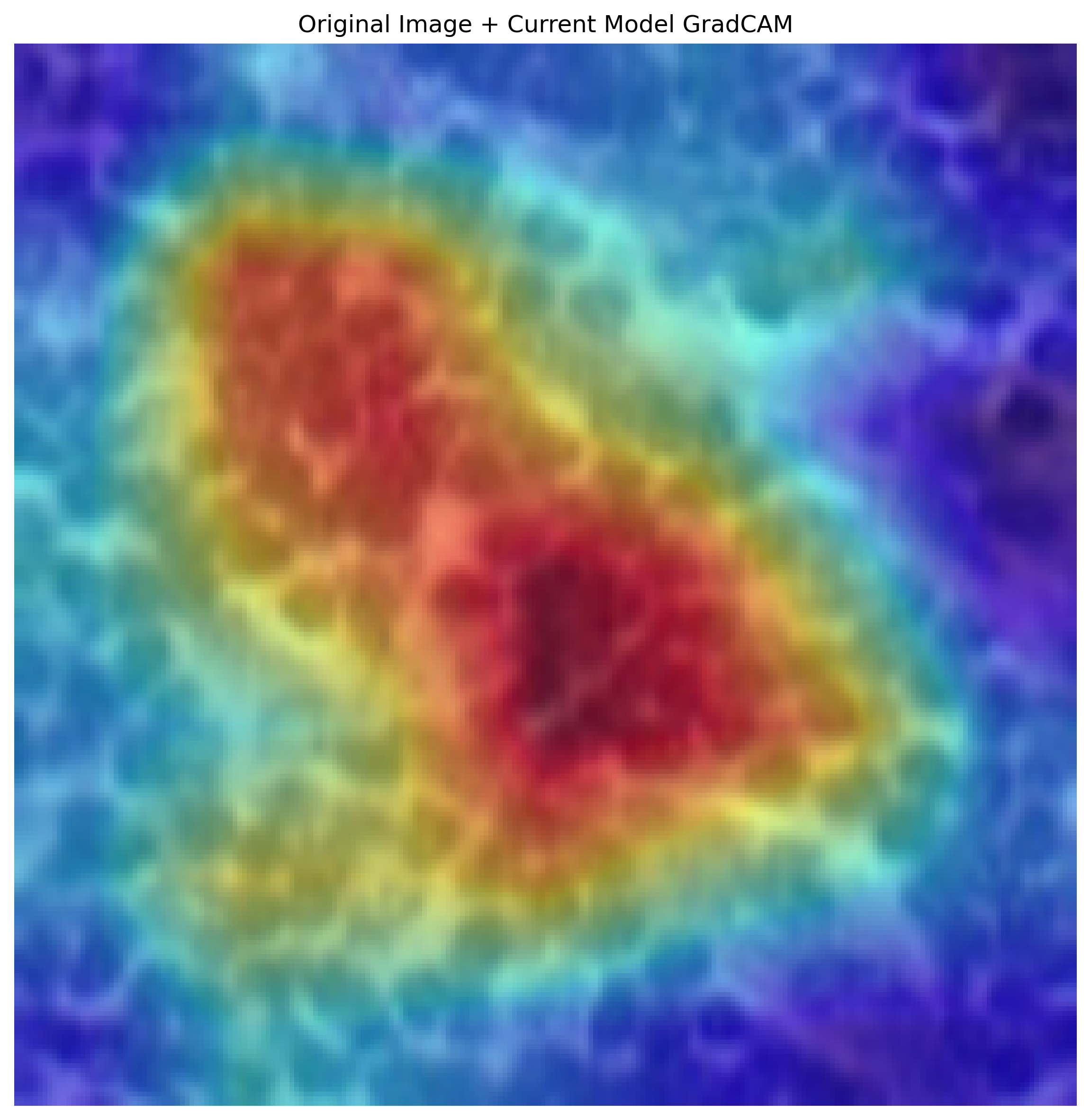} &
    \includegraphics[width=\picW]{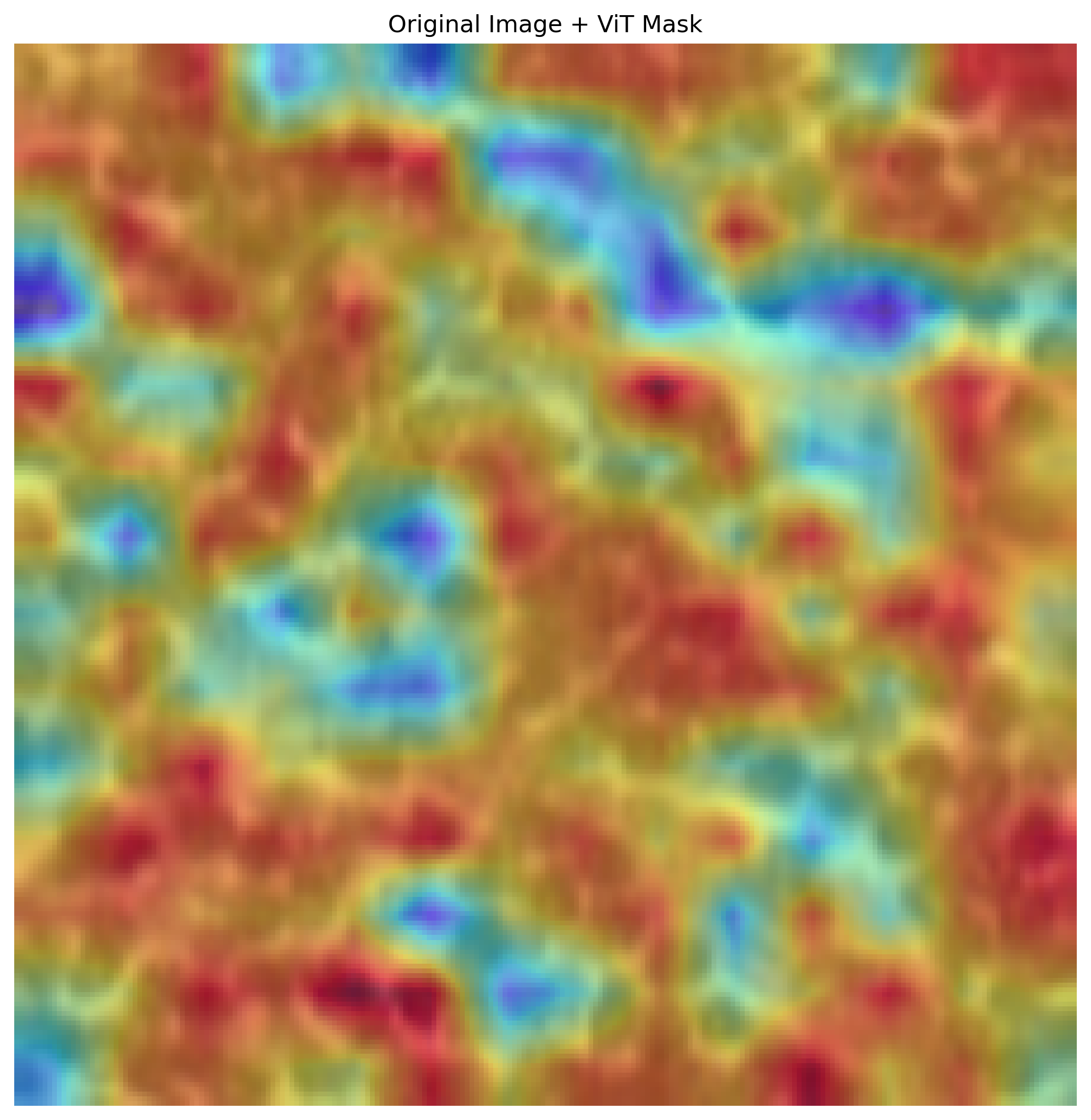} \\[-0.5ex]

    \includegraphics[width=\picW]{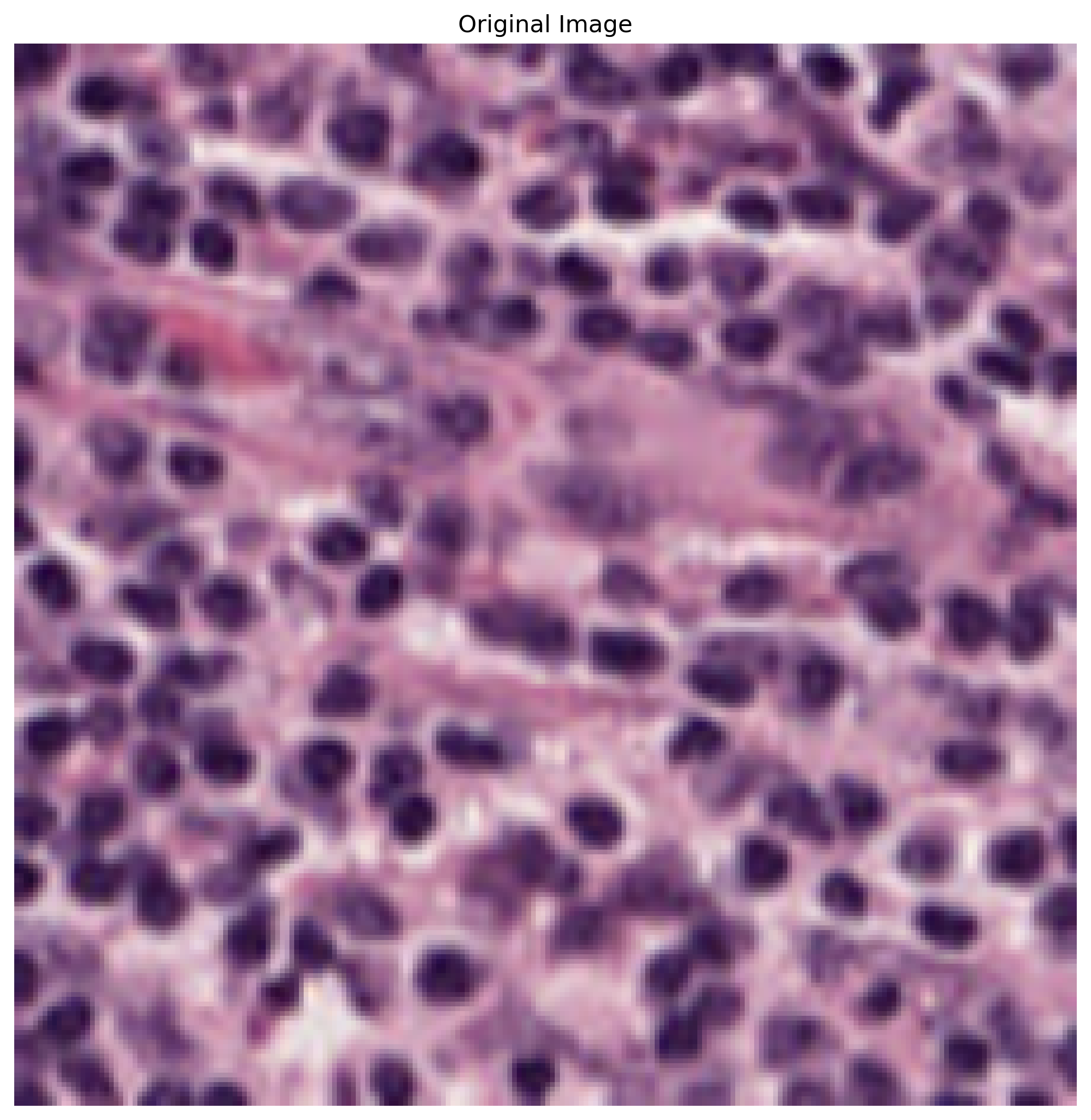} &
    \includegraphics[width=\picW]{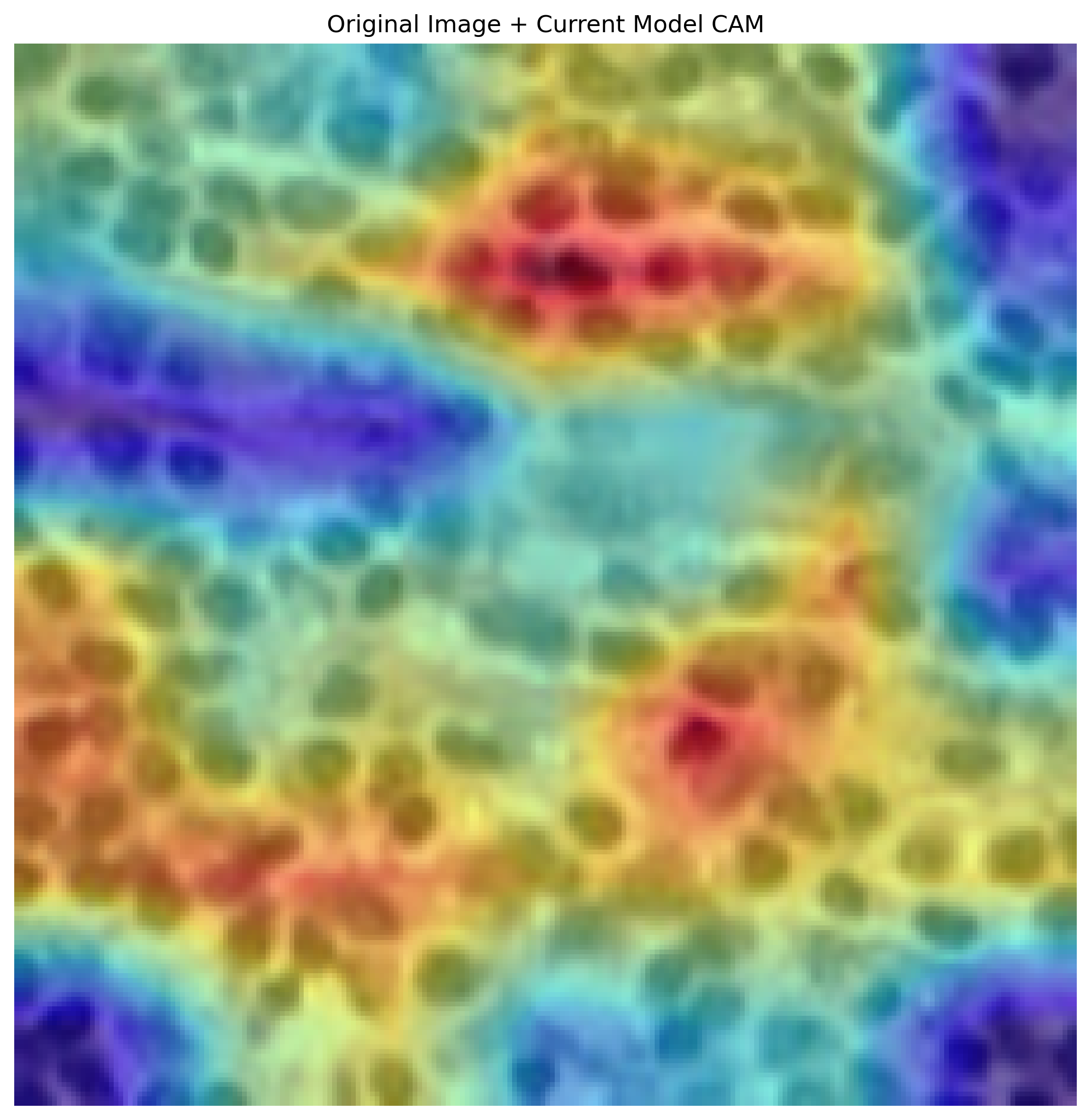} &
    \includegraphics[width=\picW]{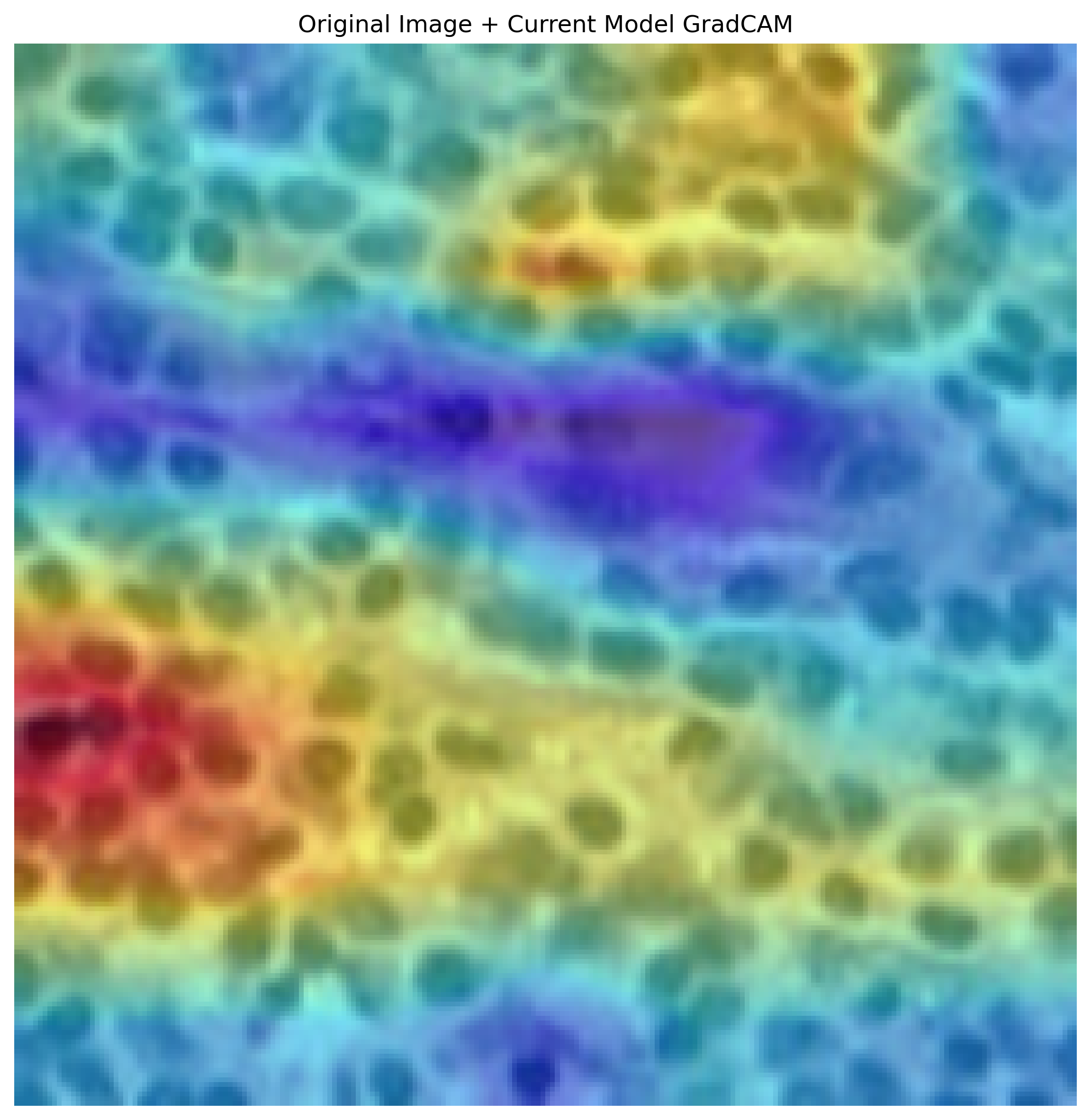} &
    \includegraphics[width=\picW]{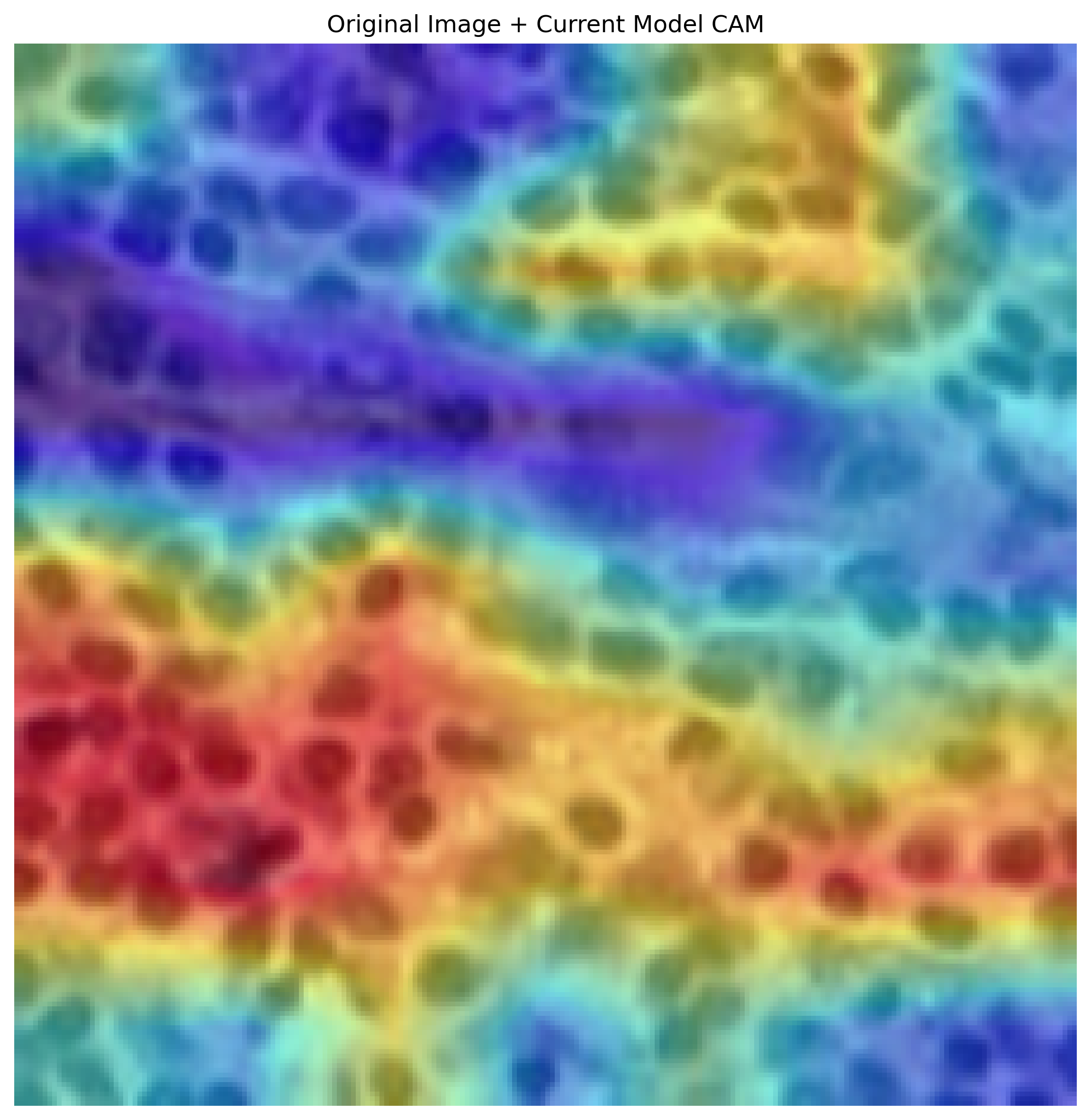} &
    \includegraphics[width=\picW]{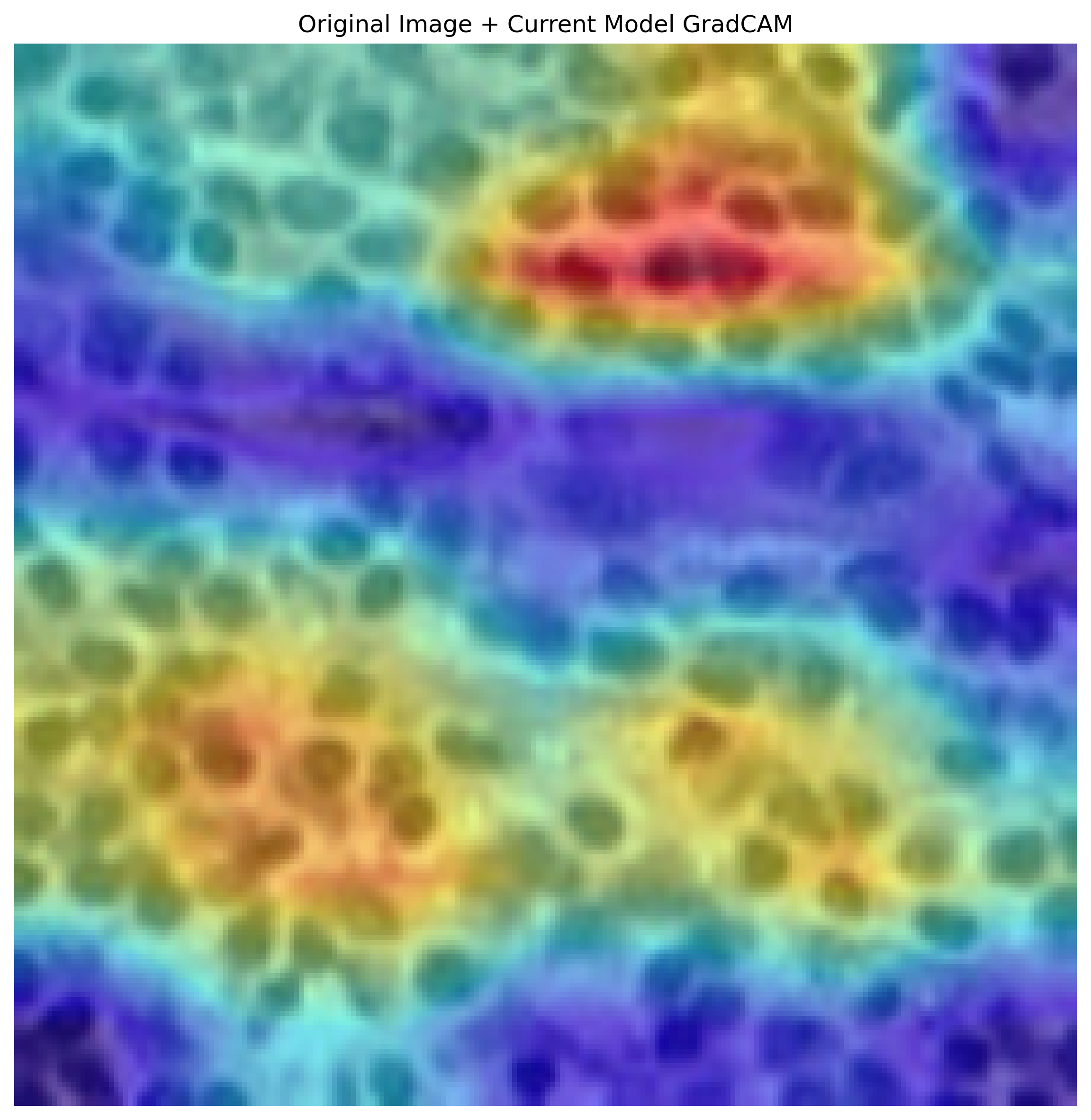} &
    \includegraphics[width=\picW]{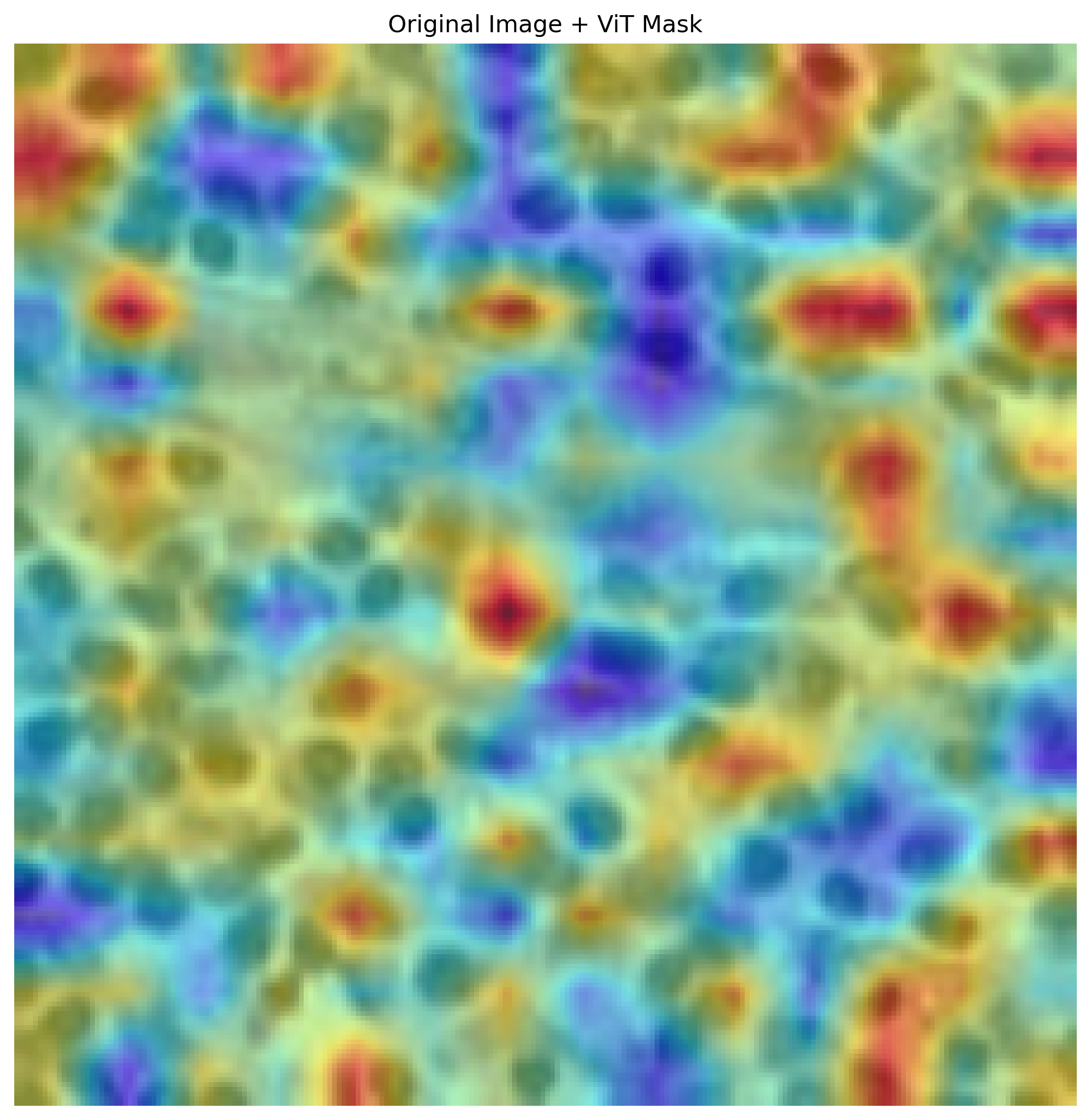} \\
  \end{tabular}

  \vskip 2pt
  \begin{tabular}{@{} *6{c} @{}}
    \makebox[\picW][c]{\subfloat[]{}} &
    \makebox[\picW][c]{\subfloat[]{}} &
    \makebox[\picW][c]{\subfloat[]{}} &
    \makebox[\picW][c]{\subfloat[]{}} &
    \makebox[\picW][c]{\subfloat[]{}} &
    \makebox[\picW][c]{\subfloat[]{}} \\
  \end{tabular}

  \caption{This figure shows the visualizations of representative samples from the training set, ID test set, and OOD test set (from top to bottom). Each row corresponds to one sample randomly selected from each respective dataset. The columns represent different visualization methods: (a) original image, (b) standard CAM, (c) standard Grad-CAM, (d) CAM generated by our CNN-based model, (e) Grad-CAM generated by our CNN-based model, and (f) foreground mask generated by our ViT-based model.}
  \label{fig:results}
\end{figure*}

\section{RESULTS AND ANALYSIS}

\begin{table}[H]
\centering
\caption{Comparison with recent domain adaptive methods on the Camelyon17 dataset. The table reports both in-domain (ID) and out-of-domain (OOD) accuracy. Results for the comparative methods are cited from~\cite{chen2023understanding}.}
\begin{tabular}{|l|l|c|c|}
\hline
INIT. & METHOD & ID Acc & OOD Acc \\
\hline\hline
ERM & IRM & 84.25(±2.2)\% & 75.68(±7.4)\% \\
ERM & IRMX & 84.54(±3.1)\% & 73.49(±9.3)\% \\
ERM & GroupDRO & 85.62(±2.4)\% & 76.09(±6.5)\% \\
ERM & VRex & 83.30(±2.5)\% & 71.06(±7.9)\% \\
\cdashline{1-4}
FeAT & IRM & 80.81(±2.8)\% & 77.97(±3.1)\% \\
FeAT & IRMX & 82.77(±1.7)\% & 76.91(±6.7)\% \\
FeAT & GroupDRO & 82.34(±1.2)\% & 80.41(±3.3)\% \\
FeAT & VRex & 82.73(±1.6)\% & 75.12(±6.6)\% \\
\cdashline{1-4}
CGP & IRM & 81.70(±2.5)\% & 81.22(±3.2)\% \\
CGP & IRMX & 81.68(±2.5)\% & 80.68(±1.6)\% \\
CGP & GroupDRO & 82.06(±2.7)\% & 81.94(±3.5)\% \\
CGP & VRex & 81.70(±2.5)\% & 81.4(±3.0)\% \\
\hline
\end{tabular}

\label{tab:camelyon17_results}
\end{table}

Table ~\ref{tab:camelyon17_results} summarizes the ID and OOD classification accuracies on the Camelyon17 dataset. Our proposed method, CGP module, is compared against two representative baselines—ERM and FeAT—under various initialization strategies, including IRM, IRMX, GroupDRO, and VRex.

In terms of ID performance, ERM achieves the highest accuracy overall, with a peak of 85.62\% under GroupDRO initialization. FeAT also performs reasonably well, reaching up to 82.77\%. CGP, while slightly behind FeAT in ID accuracy, maintains stable and competitive performance across all initializations, ranging from 81.68\% to 82.06\%.

In contrast, CGP significantly outperforms all baselines in OOD generalization. It consistently achieves the highest OOD accuracy across all initialization strategies, ranging from 80.68\% to 81.94\%, with low standard deviations (1.6\%–3.5\%). Meanwhile, ERM and FeAT show less consistent OOD performance, with larger fluctuations and standard deviations exceeding 6\%–9\% in several cases. The best OOD accuracy among baselines is 80.41\% (FeAT + GroupDRO), still below CGP's worst-case performance.

This contrast reveals a critical trade-off: CGP achieves a substantial improvement in OOD accuracy at the cost of only a minor drop in ID performance. This observation supports the findings of Teney \etal~ \cite{teney2023id}, which suggest that enhancing OOD generalization often entails a small sacrifice in ID accuracy. Our results reinforce this conclusion by demonstrating that even a slight reduction in ID accuracy (\eg from 82.77\% to 82.06\%) can result in noticeably better and more stable OOD performance.

Moreover, CGP exhibits strong robustness to the choice of initialization strategy. Its OOD performance remains consistently high and stable regardless of the optimization configuration, indicating that our method effectively mitigates the effects of distribution shift without relying on specific initialization schemes. This robustness makes CGP a promising and deployable solution for real-world medical imaging applications where data distributions are often non-stationary.

The CAM and Grad-CAM visualizations shown in Figure 3 provide compelling evidence that our proposed model more effectively localizes biologically relevant and diagnostically critical regions associated with cancerous tissues compared to current state-of-the-art approaches. Specifically, our model highlights regions characterized by increased cellular density, irregular nuclear morphology, and intensified nuclear staining—hallmarks commonly recognized in pathological assessment as closely linked to malignant transformation. These discriminative regions are crucial for accurate cancer detection and demonstrate the model’s ability to focus on meaningful histopathological features rather than superficial or confounding cues.

In contrast, our model systematically avoids regions less relevant to cancer identification, such as the cytoplasm, extracellular matrix, tissue gaps, and artifacts introduced by staining procedures. This selective attention indicates a more refined feature extraction process that prioritizes biologically significant signals over noise or technical variability, thereby enhancing model interpretability and robustness.

Furthermore, the masks generated by the ViT component corroborate this focused attention, as they similarly emphasize key pathological regions. The consistency between CNN-based Grad-CAM and ViT-based masks reinforces the reliability of our model’s interpretative capabilities across different architectural paradigms.

However, it is important to acknowledge an intrinsic limitation shared by both CNN- and ViT-based saliency methods: the necessity to upsample or interpolate relatively low-resolution masks back to the original input image dimensions. This step, while essential for visualization, inevitably imposes constraints on spatial resolution and the granularity of the highlighted regions. As a result, some fine details in the saliency maps may be smoothed or less precisely delineated, which calls for further methodological advancements to improve the spatial fidelity of such explainability techniques.

Overall, these visualization results not only validate the effectiveness of our model in capturing critical cancer-related features but also highlight areas for future improvement in interpretability and mask resolution.

\section{Conclusions}
We propose a novel intervention-based framework, CGP, to improve OOD generalization by encouraging models to rely on causally stable features. Our method leverages Vision Transformers to generate spatially fine-grained masks that guide pixel-level Gaussian perturbations, simulating causal interventions on non-essential regions. Without requiring explicit causal graph modeling, CGP effectively suppresses spurious correlations in training data. Experiments on biomedical benchmarks, CAMELYON17, demonstrate consistent performance gains, particularly on CAMELYON17, where causal cues are more visually distinct. While preliminary visualizations suggest improved model focus on causal regions, interpretability remains limited due to the lack of domain-specific knowledge.

Nonetheless, limitations remain: the quality of the ViT-generated masks can vary across datasets, and more medical benchmarks can be validated.  Future work will explore the proposed principled causal modeling methods across more domains.

\section{Acknowledgment}

The work was supported by the National Key Research and Development Program of China (Grant No. 2023YFC3306401). This research was also supported by the Zhejiang Provincial Natural Science Foundation of China under Grant No. LD24F020007, Beijing Natural Science Foundation L244043, National Natural Science Foundation of China under Grant NO. 6255000184, ``One Thousand Plan'' projects in Jiangxi Province Jxsq2023102268, Beijing Municipal Science $\&$ Technology Commission, Administrative Commission of Zhongguancun Science Park Grant No.Z231100005923035, Taiyuan City ``Double hundred Research action''  2024TYJB0127.

\bibliography{egbib}
\bibliographystyle{IEEEtran}

\end{document}